\pdfoutput=1

\documentclass[11pt]{article}

\usepackage[]{acl}

\usepackage{times}
\usepackage{graphicx}
\usepackage{latexsym}

\usepackage[T1]{fontenc}

\usepackage[utf8]{inputenc}

\usepackage{microtype}

\newcommand{\name}{\texttt{NeoDictaBERT}}
\newcommand{\biname}{\texttt{NeoDictaBERT-bilingual}}
\newcommand{\namebi}{\biname}

\title{\name: Pushing the Frontier of BERT models for Hebrew}

\author{Shaltiel Shmidman\textsuperscript{1,†}, Avi Shmidman\textsuperscript{1,2,‡}, Moshe Koppel\textsuperscript{1,2,†} \\
\textsuperscript{1}DICTA / Jerusalem, Israel \quad
\textsuperscript{2}Bar Ilan University / Ramat Gan, Israel \\ 
\texttt{\small \textsuperscript{†}\{shaltieltzion,moishk\}@gmail.com} \\
\texttt{\small \textsuperscript{‡}avi.shmidman@biu.ac.il}}

\begin{document}
\maketitle

\begin{abstract}
Since their initial release, BERT models have demonstrated exceptional performance on a variety of tasks, despite their relatively small size (BERT-base has \textasciitilde110M parameters). Nevertheless, the architectural choices used in these models are outdated compared to newer transformer-based models such as Llama3 and Qwen3. In recent months, several architectures have been proposed to close this gap. ModernBERT and NeoBERT both show strong improvements on English benchmarks and significantly extend the supported context window. Following their successes, we introduce {\name} and {\namebi}: BERT-style models trained using the same architecture as NeoBERT, with a dedicated focus on Hebrew texts. These models outperform existing ones on almost all Hebrew benchmarks and provide a strong foundation for downstream tasks. Notably, the \namebi model shows strong results on retrieval tasks, outperforming other multilingual models of similar size. In this paper, we describe the training process and report results across various benchmarks. We release the models\footnote{The models are released under a CC BY-SA license: \url{https://creativecommons.org/licenses/by-sa/4.0/}} to the community as part of our goal to advance research and development in Hebrew NLP.
\end{abstract}

\section{Introduction}

Transformer-based encoder models first achieved popularity with the release of the original BERT models \cite{devlin2019bert}. Since then, multiple models have been released based on the original BERT, incorporating various modifications to boost performance. Examples include RoBERTa \cite{liu2019roberta}, DeBERTa \cite{he2021deberta}, ALBERT \cite{lan2020albert}, and CharacterBERT \cite{boukkouri2020characterbertreconcilingelmobert}.

In recent years, transformer-based decoder models such as GPT-3 \cite{brown2020languagemodelsfewshotlearners} and Llama \cite{touvron2023llamaopenefficientfoundation} have risen in popularity, driving extensive research into architectural choices of transformer-based models. As a result, many changes to the original transformer architecture have become mainstream, aimed at improving scalability, extending context windows, accelerating convergence, and enhancing performance.

Yet, while decoder models steadily evolved, encoder models largely retained their original architecture. This gap led to the recent release of ModernBERT \cite{warner-etal-2025-smarter} and subsequently NeoBERT \cite{breton2025neobertnextgenerationbert}. The key motivation behind these releases was to modernize the BERT-style architecture using innovations from newer decoder models, scaling up pretraining data by an order of magnitude. Both demonstrated substantial performance improvements on downstream tasks, with NeoBERT outperforming ModernBERT on several key benchmarks.

In this paper, we present our new models for Hebrew NLP, {\name} and {\biname}, which adopt the NeoBERT architecture and training methodology, and which are trained on Hebrew-centric pretraining data for Hebrew language modeling. Both models natively support a context window of 4,096 tokens (8x more than the original BERT). The {\name} model was trained solely on Hebrew data, while {\biname} was trained on a 60–40 mix of English and Hebrew data, respectively. We observe performance gains across almost all benchmarks when compared with previous state-of-the-art Hebrew models; the most dramatic boost occurs in QA tasks, demonstrating significant improvements in semantic understanding. Additionally, we show that {\biname} performs strongly when fine-tuned for retrieval, on both English and Hebrew benchmarks.

\section{Approach}
\subsection{Architecture}

As noted, we adopt the NeoBERT architecture, which introduces several key changes compared to the original BERT. In particular:

\begin{itemize}
\item \textbf{Normalization}: RMSNorm \cite{zhang2019rmsnorm} applied before the main sub-layer (Pre-RMSNorm), instead of the standard LayerNorm applied after the main sub-layer and residual (Post-LayerNorm).
\item \textbf{Position Encoding}: RoPE \cite{su2023roformerenhancedtransformerrotary}, which encodes relative positions, allowing better generalization to longer sequences.
\item \textbf{Activation Function}: SwiGLU gated activation \cite{shazeer2020gluvariantsimprovetransformer}, replacing the non-gated GeLU \cite{hendrycks2023gaussianerrorlinearunits} activation.
\item \textbf{Shared Embedding \& Output Weights}: The embedding layer and output LM-head weights are no longer shared.
\end{itemize}

Two additional architectural changes are worth noting. First, the context window: while the original BERT models are trained initially with a 128-token window, and then extended to 512 tokens in phase 2 of the training, here we train with an initial window of 1,024 tokens, extended to 4,096 tokens in the second phase. Second, we adjust the depth-to-width ratio, increasing the model depth to 28 layers (compared to 12 in $BERT_{base}$) while retaining the original $BERT_{base}$ width of 768. This achieves the theoretically optimal ratio as suggested by \citet{levine2020limits}.

\begin{table*}[ht]
\centering
\renewcommand{\arraystretch}{1.2}
\setlength{\tabcolsep}{6pt}
\begin{tabular}{lcccccccc}
\hline
\textbf{Model} & \multicolumn{3}{c}{\textbf{QA}} & \multicolumn{2}{c}{\textbf{Dependency}} & \textbf{Sentiment} & \textbf{NER} \\
 & EM & F1 & T1nls & UAS & LAS & Acc & F1 \\
\hline
\texttt{mBERT} & 69.08 & 74.32 & 77.01 & -- & -- & 84.21 & 79.11 \\
\texttt{AlephBertGimmel} & 61.77 & 67.97 & 71.35 & -- & -- & 89.51 & 85.26 \\
\texttt{DictaBERT} & 70.35 & 77.04 & 80.34 & 90.52 & 86.54 & \textbf{89.79} & 87.01 \\
\texttt{mmBERT-base} & 72.14 & 77.75 & 80.7 & 91.63 & 87.9 & 87.32 & 72.49 \\
\name & 76.40 & 82.83 & 85.86 & 91.77 & 87.80 & 89.61 & 86.97 \\
\biname & \textbf{82.30} & \textbf{87.49} & \textbf{90.38} & \textbf{92.32} & \textbf{88.90} & 89.62 & \textbf{90.43} \\
\hline
\end{tabular}
\caption{Performance on Hebrew Language benchmarks}
\label{tab:he-benchmarks}
\end{table*}

\begin{table*}[ht!]
\centering
\renewcommand{\arraystretch}{1.05}
\setlength{\tabcolsep}{3pt}
\begin{tabular}{lcccccccc}
\hline
\textbf{Model Name} & \textbf{Pair Class.} & \textbf{Class.} & \textbf{STS} & \textbf{Retrieve} & \textbf{Cluster} & \textbf{Rerank} & \textbf{Summ.} & \textbf{Avg} \\
\hline
\texttt{ModernBERT-base} & 80.5 & 65.6 & 75.5 & 44.8 & 43.3 & 44.1 & 22.6 & 53.8 \\
\texttt{mmBERT-base} & 80.2 & 64.8 & 74.8 & 44.9 & 41.7 & 44.9 & 26.0 & 53.9 \\
\namebi & 82.2 & 67.4 & 74.4 & 44.4 & 37.1 & 42.8 & 24.5 & 53.2 \\
\hline
\end{tabular}
\caption{Performance on MTEB v2 English}
\label{tab:mteb-v2-eng}
\end{table*}

\subsection{Tokenizer}

Both models use the WordPiece tokenization method proposed by \newcite{song2021fast}, with the default normalizers and preprocessors suggested by HuggingFace, along with Hebrew-specific modifications introduced in DictaBERT \cite{shmidman2023dictabertstateoftheartbertsuite}.

The {\name} model uses the same tokenizer as DictaBERT, while for {\biname} we train a new tokenizer on a mixture of Hebrew and English, maintaining a vocabulary size of 128,000.
\subsection{Data}

\begin{itemize}

\item \textbf{English} We use FineWeb-Edu \cite{lozhkov2024fineweb-edu}, an open-source, high-quality pre-training dataset for English.

\item \textbf{Hebrew} We begin with the training corpus used in \citet{shmidman2024adaptingllmshebrewunveiling}, and we expand it to include new Hebrew corpora which have become available since.

\end{itemize}

\subsection{Training Details and Hyperparameters}

Following the NeoBERT training recipe, we train our model only on the masked-language modeling task (and not on the next-sentence prediction task). In addition, we train on $20\%$ of the tokens (as opposed to $15\%$) and mask the token $100\%$ of the time (as opposed to splitting between masking, swapping to an incorrect token, and leaving the token unchanged).

Both models were trained with a batch size of 2048, using packed sequences to minimize padding and maximize data utilization. The initial training phase used a context window of 1,024 tokens and a maximal learning rate of 6e-3, followed by a second phase extending the context window to 4,096 tokens with a maximal learning rate of 1e-4. For both phases, we used the AdamW optimizer \cite{loshchilov2018decoupled} with 500 warmup steps and cosine annealing. The models were trained using the NeMo Framework mixed-FP8 precision plugin, resulting in a 30–50\% boost in training throughput.

The models were trained using the NeMo Framework \cite{Harper_NeMo_a_toolkit}, a scalable generative AI framework built for researchers and developers working on large language models, optimized for large-scale model training on NVIDIA hardware..

\textbf{\name} was pre-trained on 8×H100 80GB GPUs. The first phase trained on 232B tokens (5 epochs on our Hebrew data) for a total of 88 hours. The second phase trained on 46B tokens (1 epoch) for a total of 23 hours.

\textbf{\biname} was pre-trained on 16×H200 141GB GPUs on NVIDIA DGX Cloud Lepton. The first phase trained on 612B tokens (60-40 English-Hebrew split) for a total of 106 hours. The second phase trained on 122B tokens for a total of 28 hours.

\section{Hebrew Experiments and Results}

We compare the performance of {\name} and {\biname} with previous Hebrew models of similar parameter size, drawing results from their respective publications when available; otherwise, we fill in the scores by fine-tuning the models ourselves. The models we compare to are mBERT \cite{devlin2019bert}, AlephBertGimmel \cite{gueta2023large}, DictaBERT \cite{shmidman2023dictabertstateoftheartbertsuite,shmidman-etal-2024-mrl}, and mmBERT \cite{marone2025mmbertmodernmultilingualencoder} (multilingual-ModernBERT).

Following previous publications, we tested our models on the following Hebrew benchmarks:

\textbf{Dependency Parsing} We follow the experimental setup of \citet{shmidman-etal-2024-mrl}, evaluating both UAS (unlabeled accuracy score) and LAS (labeled accuracy score) on whole words in Hebrew. We use the Hebrew UD Treebank introduced by \citet{tsarfaty2013unified} and revised to match the tagging schema of IAHLT \cite{ZeldesHowellOrdanBenMoshe2022}.\footnote{The dataset is publicly released and available at \url{https://github.com/IAHLT/UD_Hebrew}}

\textbf{Named Entity Recognition (NER)} We train on the NEMO dataset presented by \citet{barekettsarfatinemo}, which contains nine categories and 6,220 sentences (7,713 entities). We report the token-level F1 score for each model.

\textbf{Sentiment Analysis} We evaluate our models on the sentiment analysis dataset presented by \citet{amram-etal-2018-representations}, based on 12K social media comments.

\textbf{Question Answering} For this task, we use the HeQ dataset by \citet{Cohen2023HeQ}, which contains 30K high-quality samples from the GeekTime newsfeed and Wikipedia.

The results for these tasks are displayed in Table \ref{tab:he-benchmarks}. As can be seen, {\name} and {\biname} outperform all previous models, with a noticeable improvement in QA scores, indicating a deeper semantic understanding.

\section{Retrieval Experiments and Results}

We also evaluate the retrieval performance of {\biname}. We follow the training setup used by \citet{marone2025mmbertmodernmultilingualencoder} and fine-tune the model for retrieval using English-only datasets.

We compare the performance of the trained model on the MTEB v2 English benchmarks \cite{muennighoff-etal-2023-mteb}; results are shown in Table \ref{tab:mteb-v2-eng}. The model demonstrates performance comparable to other leading models, underscoring its strength.

We further evaluate the trained model on a Hebrew-only retrieval benchmark by submitting it to the \texttt{Hebrew Semantic Retrieval National Challenge}.\footnote{\url{https://www.codabench.org/competitions/9950/}} We use the same code and baseline setup, replacing the embedding model (multilingual-e5-base \cite{wang2024multilingual}) with our fine-tuned model. Notably, even though our model was fine-tuned only on \textit{English} data, it improves the baseline NDCG@20 score from 0.397 to 0.404, reflecting effective generalization from the English training corpus to Hebrew input texts.

\section{Conclusion}

We are happy to release these models to the public to help advance research and development in Hebrew NLP.\footnote{{\name} is available at \url{https://huggingface.co/dicta-il/neodictabert} \ {\biname} is available at \url{https://huggingface.co/dicta-il/neodictabert-bilingual}}

\section{Acknowledgments}

We would like to thank the NVIDIA DGX Cloud Lepton team for early access to the platform, and for providing us with the necessary compute to train this model. 

\bibliography{anthology}
\bibliographystyle{acl_natbib}

\end{document}